\documentclass[sigconf]{acmart}
\usepackage{pdfpages}
\usepackage{arydshln}
\usepackage{diagbox}
\usepackage{CJKutf8}
\usepackage[normalem]{ulem}

\usepackage{amsmath}
\usepackage{caption}
\usepackage{booktabs}

\usepackage{marvosym}

\usepackage{enumitem}
\usepackage{graphicx}
\usepackage{multirow}
\usepackage{makecell}
\usepackage{color}
\usepackage{bm}

\usepackage{bbding}

\newcommand{\eg}{\textit{e.g.}}
\newcommand{\ie}{\textit{i.e.}}
\newcommand\ul{\underline}
\theoremstyle{remark}

\AtBeginDocument{%
  \providecommand\BibTeX{{%
    \normalfont B\kern-0.5em{\scshape i\kern-0.25em b}\kern-0.8em\TeX}}}

\copyrightyear{2024}
\acmYear{2024}
\acmConference[MM '24]{Proceedings of the 32nd ACM International Conference on Multimedia}{October 28-November 1, 2024}{Melbourne, VIC, Australia} 
\acmBooktitle{Proceedings of the 32nd ACM International Conference on Multimedia (MM '24), October 28-November 1, 2024, Melbourne, VIC, Australia}

\begin{document}

\title{See or Guess: Counterfactually Regularized Image Captioning}

\author{
Qian Cao\textsuperscript{\textmd{1}}, Xu Chen\textsuperscript{\textmd{1}\Letter}, Ruihua Song\textsuperscript{\textmd{1}\Letter}, Xiting Wang\textsuperscript{\textmd{1}\Letter}, Xinting Huang\textsuperscript{\textmd{2}}, Yuchen Ren\textsuperscript{\textmd{1}}
}

\email{{caoqian4real, xu.chen, rsong, xitingwang, siriusren}@ruc.edu.cn, timxinhuang@tencent.com}
\affiliation{
\institution{\textsuperscript{\textmd{1}}Gaoling School of Artificial Intelligence, Renmin University of China}
\institution{\textsuperscript{\textmd{2}}Tencent AI Lab}
\country{}
}
   
\thanks{\Letter\ Corresponding author.}

\renewcommand{\authors}{Qian Cao, Xu Chen, Ruihua Song, Xiting Wang, Xinting Huang, Yuchen Ren}
\renewcommand{\shortauthors}{Qian Cao et al.}


\begin{abstract}
Image captioning, which generates natural language descriptions of the visual information in an image, is a crucial task in vision-language research.
Previous models have typically addressed this task by aligning the generative capabilities of machines with human intelligence through statistical fitting of existing datasets.
While effective for normal images, they may struggle to accurately describe those where certain parts of the image are obscured or edited, unlike humans who excel in such cases.
These weaknesses they exhibit, including hallucinations and limited interpretability, often hinder performance in scenarios with shifted association patterns.
In this paper, we present a generic image captioning framework that employs causal inference to make existing models more capable of interventional tasks, and counterfactually explainable.
Our approach includes two variants leveraging either total effect or natural direct effect.
Integrating them into the training process enables models to handle counterfactual scenarios, increasing their generalizability.
Extensive experiments on various datasets show that our method effectively reduces hallucinations and improves the model's faithfulness to images, demonstrating high portability across both small-scale and large-scale image-to-text models.
The code is available at \url{https://github.com/Aman-4-Real/See-or-Guess}.
\end{abstract}

\keywords{Image Captioning, Counterfactual Causal Inference, Object Hallucination, Image-to-text Generation}

\maketitle

\begin{figure}[t!]
\includegraphics[width=1\linewidth]{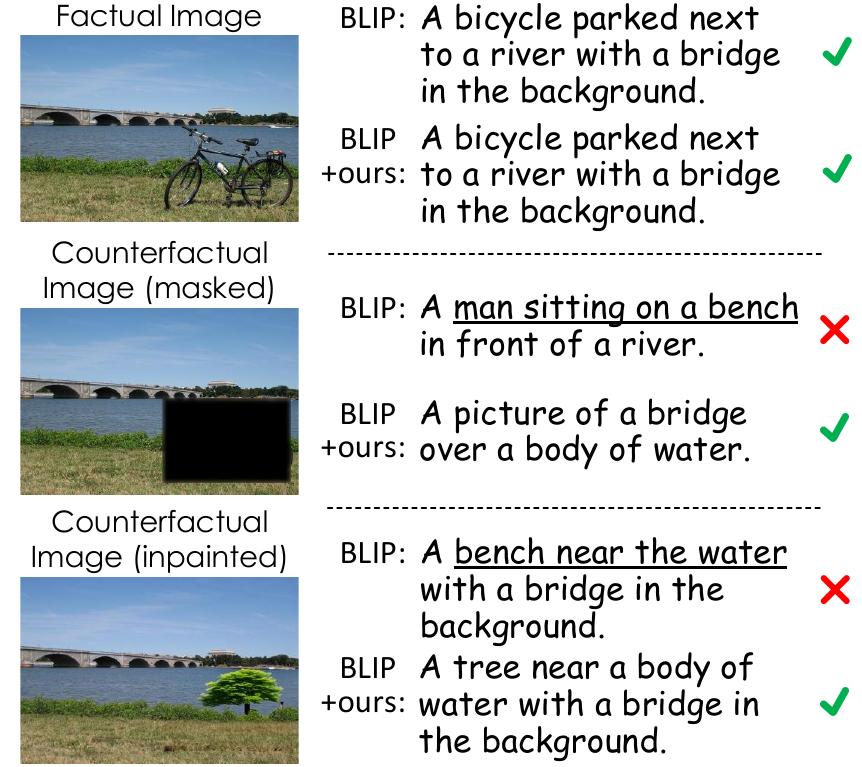}
\caption{An example of generated captions of different methods in the factual and two counterfactual scenarios.}
\label{fig:teaser}
\end{figure}

\section{Introduction}
As a fundamental task in vision-language understanding research, image captioning requires models to mimic the human ability to compress huge amounts of visual information into descriptive language~\cite{mokady2021clipcap,xu2015show,Barraco_2023_ICCV}.
A large amount of image-to-text methods~\cite{li2022blip,cao2022multi,stefanini2022show} have been developed, among which recent large multimodal models~\cite{li2023blip2,llava,minigpt4} perform surprisingly well in describing an image in details.
Despite their good performance in real scenarios, their capabilities still differ from those of humans in interventional scenarios.
For example, in Figure~\ref{fig:teaser}, the BLIP model~\cite{li2022blip} can generate a sentence that accurately describes the factual image.
However, when the bicycle is masked or changed to a tree as shown in the counterfactual images, it generates incorrect descriptions such as ``A man sitting on a bench in front of a river.''
Such errors reveal that the model might not have precisely understood the image. Instead, it may make guesses based on common association patterns in the datasets.
For example, the frequent co-occurrence of a river and a man in the dataset may lead the model to form shortcut connections and wrongly generate ``man'' for most images with a river.

The above analysis suggests that while current models may exhibit impressive performance, it does not necessarily imply their ability to accurately comprehend the contents of an image and generate appropriate descriptions, a capability inherently possessed by humans.
Such weaknesses may result in hallucinations and hinder the interpretability of models since people cannot exactly tell which parts of the image correspond to the generated words in the text. 
Furthermore, when these learned models are applied to other scenarios with shifted association patterns, their performance may suffer a substantial deterioration.

To overcome these shortcomings, we design a novel framework that integrates causal inference into any image captioning model to mitigate shortcut correlations.
Specifically, we utilize counterfactual concepts to enhance the correspondence between visual and textual characteristics. 
The core idea is that when certain regions of an image are removed, the generated text should not describe those regions.
While this idea is intuitive, it is challenging to implement for the following reasons.
First, existing counterfactual models primarily focus on classification tasks~\cite{abbasnejad2020counterfactual,goyal2019counterfactual,yue2021counterfactual}. 
However, we handle a generation task that necessitates the consideration of sequential impacts between words.
In our multi-modal scenario, the influence of the image on the word can be attributed to two paths: (1) a direct influence from the image to the word, and (2) an indirect influence, where the image first impacts preceding words, which subsequently influence the current word.
It presents a nontrivial challenge to distinguish between these two paths and enhance the first one to minimize hallucination while preserving linguistic fidelity.\looseness=-1

To address the above challenges, in this paper, we first formalize the image captioning task as a sequential causal graph,
where each word in the generated text is influenced by both its preceding words and the image. 
Following this causal graph, we leverage the causal concepts of total effect (TE) and natural direct effect (NDE) to distinguish the different reasons behind word generation.
Then we can intervene in the cause, and enhance the correspondences between the image and words while controlling the other influential factors. 
Finally, we propose a counterfactually regularized image captioning framework.
Our main contributions are as follows:

$\bullet$ We propose a generic framework to counterfactually regularize image captioning models and thus make them more human-like, explainable, and robust.

$\bullet$ We propose two causal methods based on total effect and natural direct effect to enhance the correspondence between the visual and textual characteristics.

$\bullet$ Extensive experiments on various models and datasets demonstrate the high generality and interpretability of our methods, which can effectively reduce object hallucinations and enhance model faithfulness to the images.

\section{Related Work}
\subsection{Image Captioning}
Image captioning, crucial for image-to-text generation~\cite{stefanini2022show}, has evolved from convolution neural network (CNN)-based encoders and recurrent neural network (RNN)-based decoders~\cite{vinyals2015show,yao2018exploring} to Transformer architectures~\cite{li2019entangled,cornia2020meshed}, and further into vision language pretraining (VLP) models~\cite{li2020oscar,zhou2020unified,li2022blip}.
Recent advancements in visual pertaining~\cite{santurkar2023caption,wang2023densecl} and Large Vision Language Models~\cite{li2023blip2,llava,minigpt4} have sparked renewed interest in the field.
In addition, some works explore integrating multimodal representation models like CLIP~\cite{radford2021learning} to furnish visual support for language models~\cite{mokady2021clipcap}.
However, our approach has a model-agnostic nature and flexibility.
Due to the notable performance of VLP models, we validate the effectiveness across various architectures (decoder-only and encoder-decoder) and model scales by employing ClipCap~\cite{mokady2021clipcap}, BLIP~\cite{li2022blip}, and BLIP2~\cite{li2023blip2} as backbones models.

\subsection{Object Hallucination in Image Captioning}
Alleviating hallucination of image captioning models does not solely hinge on improved image perception capability but also on factors like over-reliance on language priors or biases during sequence generation~\cite{rohrbach2018object,Petryk_2024_WACV}, potentially leading to guesswork that is not faithful to the image.
Researchers~\cite{rohrbach2018object} thus propose utilizing the CHAIR metric to quantify hallucination occurrence.
Some efforts have been made to reduce model reliance on common or biased co-occurrences by adjusting object label co-occurrence statistics~\cite{biten2022let}.
Other methods reduce object hallucinations and maintain semantic consistency by learning consensus representations through aligning scene and language graphs~\cite{zhang2021consensus}, or by aligning textual tokens and visual objects using masked language modeling~\cite{dai2022plausible}.
However, these methods may blur semantic and visual alignments and over-rely on dataset co-occurrence patterns, harming interpretability and performance in real scenarios.
While some works consider causal modeling~\cite{yang2021deconfounded,liu2022show}, they often require altering the model structures.
Our approach is more general and aims to establish the correct vision-to-language relationship during word generation.

\subsection{Counterfactual Causal Inference}
Causal inference seeks to unravel the causal relationships and underlying mechanisms driving observed outcomes~\cite{pearl2010causal,balke2013counterfactuals,glymour2016causal,wu2023causality}.
Moreover, counterfactual causal inference offers a framework to enhance~\cite{abbasnejad2020counterfactual,teney2020learning} and explain~\cite{hendricks2018generating,goyal2019counterfactual} models in counterfactual scenarios.
However, the majority of these counterfactual-related works are tailored for classification tasks, such as image classification~\cite{abbasnejad2020counterfactual,goyal2019counterfactual,yue2021counterfactual}, representations learning~\cite{teney2020learning,zhang2020counterfactual}, or visual question answering~\cite{hirota2021visual,niu2021counterfactual,liang2020learning}, rather than for generation tasks.
Classification tasks exhibit a deterministic correspondence between input and output, whereas, in the generation process, the counterfactual image and preceding generated tokens collectively influence the subsequent token generation, creating an effect propagation. 
Some researchers~\cite{wang2021counterfactual} have applied Maximum Likelihood Estimation (MLE) on interventional distributions to address spurious correlations caused by observed confounding factors.
However, the applicability of their framework is constrained by the strong ignorability assumption and lacks causal analysis in multi-modal scenarios.
Capturing this causal correspondence~\cite{pearl2022direct} is challenging, especially in multimodal scenarios, and has thus received little attention in prior literature.
In this paper, we endeavor to leverage counterfactual causal inference to tackle this challenge and gain insights into model generation behavior.

\section{Preliminaries: A Causal Look at Image Captioning}

This section presents the fundamental concepts and notations of causal inference~\cite{glymour2016causal,pearl2022direct} and how we apply it in image captioning.
In the following, capital letters, \ie, cause $X$, Mediator $M$, and Effect $Y$, represent random variables.
The values or subscripts of these random variables indicate their observed values.

As for image captioning, a model is used to process an input image $I$ and produce a corresponding textual description, \ie, a sequence $S=(s_{1},s_{2},\dots,s_{L})$
, where $s_{i}$ is a token in the sequence and $L$ is the sequence length.
The sequence of the preceding tokens of $s_{j}$ is denoted as $S_{<j}=(s_{1},s_{2},\dots,s_{j-1})$.
We will later present how to treat these variables from a causal perspective.

\subsection{Causal Graph}

\begin{figure}[t!]
\includegraphics[width=1.\linewidth]{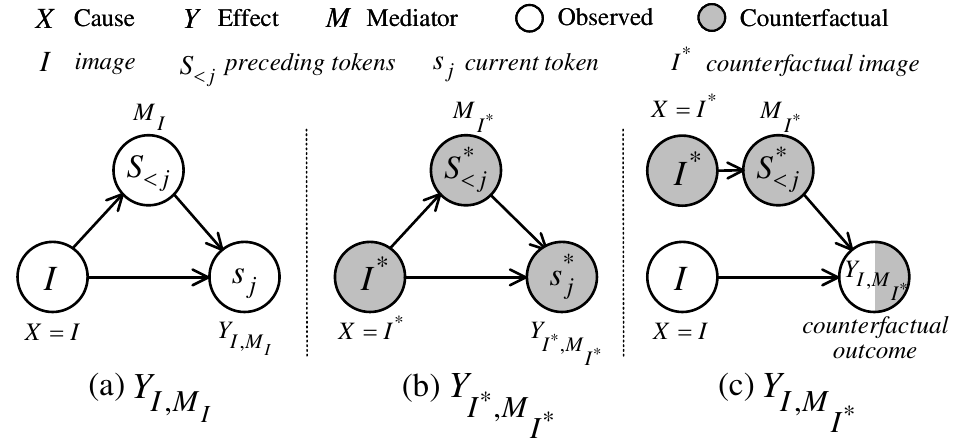}
\caption{Illustration of causal graphs and counterfactual causal effect notations.}
\label{fig:causal_graph}
\end{figure}

A causal graph describes the causal relations between different random variables in a graph manner~\cite{pearl2000models}.
In a causal graph $\mathcal{G}=\{\mathcal{V}, \mathcal{E}\}$, a node $\boldsymbol{v}\in\mathcal{V}$ represents a variable and a directed edge $\boldsymbol{e}\in\mathcal{E}$ represents a causal relationship between variables.
The \textit{direct effect} means that there is an edge between two variables, \eg, in Figure~\ref{fig:causal_graph}~(a), $X$ has a \textit{direct effect} on $Y$.
The \textit{indirect effect} means that two variables are not directly linked, but are connected via some \textit{mediator} variables, \eg, $X$ has a \textit{indirect effect} on $Y$ if $X \rightarrow M \rightarrow Y$.

Considering the process of auto-regressive generation, at each step, the current token $s_j$ is determined by all the preceding tokens $S_{<j}$, and the visual information of the input image $I$ as well.
As shown in Figure~\ref{fig:causal_graph}~(a), at step $j$, $S_{<j}$ is influenced by $I$, and $s_j$ is jointly determined by $I$ and $S_{<j}$.
We use $Y_{I,M_{I}} = Y(X=I, M=M_{I})$ to denote the probability of token $s_j$ when the cause $X$ is set to $I$ and the mediator $M$ is set to $M_{I}=S_{<j}$.

\begin{figure*}[t!]
\includegraphics[width=1.\linewidth]{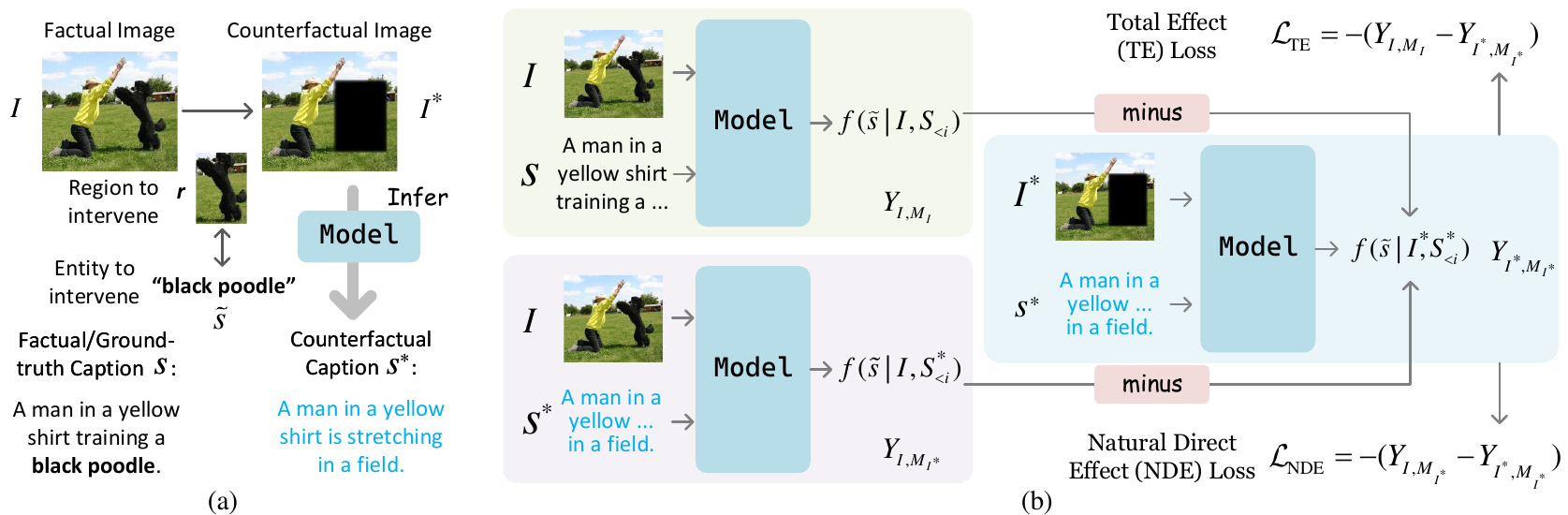}
\caption{Our framework of counterfactual regularization. (a) shows how to prepare counterfactual images and captions by example. (b) illustrates how the TE loss and NDE loss are calculated in the example. Counterfactual captions are in blue. The phrase corresponding to the image region in the mask is ``black poodle''. Best viewed in color.}
\label{fig:framework}
\end{figure*}

\subsection{Counterfactual Causal Effects}
In causal inference, counterfactual causal effects compare hypothetical outcomes under factual and counterfactual treatments~\cite{pearl2010causal,balke2013counterfactuals}.

As shown in Figure~\ref{fig:causal_graph}~(b), the value of the counterfactual of variable $X$ is equal to the counterfactual image $I^{*}$, where $I^{*}$ is created by intervening in the factual image $I$. The hypothetical outcome of $Y$ is denoted as $Y_{I^{*},M_{I^{*}}} = Y(X=I^{*}, M=M_{I^{*}})$, where the mediator $M_{I^{*}}=S_{<j}^*$.
The total effect (TE) is the difference between two hypothetical conditions: one being factual transition where $X=I$ (under treatment, corresponding to Figure~\ref{fig:causal_graph}~(a)) and the counterfactual being $X=I^{*}$ (under no-treatment, corresponding to Figure~\ref{fig:causal_graph}~(b)). Mathematically, the total effect can be expressed as
\begin{equation}
    \text{TE}_{I, I^{*}} = Y_{I,M_{I}} - Y_{I^{*},M_{I^{*}}}. \label{formula:te}
\end{equation}
$\text{TE}_{I, I^{*}}$ measures the effect of all factors (\ie, direct and indirect effects) resulting from changing image $I$ to $I^*$.\looseness=-1 

Further, intervening both $X$ and $M$ allows the total effect to be decomposed into two components, namely the natural direct effect (NDE) and the total indirect effect (TIE).
Unlike TIE which focuses on the effect brought by changes in the mediator $M$, NDE is the effect of $X$ on $Y$ that results solely from changes in $X$, without any influence from $M$, which can be denoted as
\begin{equation}
    \text{NDE}_{I, I^{*}} = Y_{I,M_{I^{*}}} - Y_{I^{*},M_{I^{*}}}. \label{formula:nde}
\end{equation}
The first term $Y_{I,M_{I^{*}}}$ corresponds to Figure~\ref{fig:causal_graph}~(c), which keeps $X=I$ and conducts intervening on $M$ via $I^*$ to form a counterfactual outcome $Y_{I,M_{I^*}}$. The second term $Y_{I^{*},M_{I^{*}}}$ corresponds to Figure~\ref{fig:causal_graph}~(b).
Formula~\ref{formula:nde} describes the variation of $Y$ when $X$ is changed from $I$ to its counterfactual $I^{*}$ while $M$ is held constant at $M(X=I^{*})$.

This paper explores how to use TE or NDE to reduce object hallucination in image captioning and improve interoperability.

\section{Counterfactual Regularization}

In this section, we detail the construction of counterfactual data, present our framework and design two counterfactual regularization losses generally applicable to existing image captioning models.

\subsection{Constructing Counterfactual Data}
\label{sec:construct_cf_data}
Collecting counterfactual images for the factual ones is challenging.
However, by adding a mask, it is easy to achieve minimal changes to the original image when constructing the counterfactual one, which can be regarded as an approximation of the idealized counterfactual image.
Specifically, we construct counterfactual images by using datasets with labeled bounding boxes for corresponding phrases in the image captions.
As shown in Figure~\ref{fig:framework}~(a), we first select the entity to intervene ($\tilde{S}$), \eg, ``black poodle''.
Then we identify its corresponding region to intervene ($r$) in the image based on the labeled bounding boxes.
A black mask is used to replace the region $r$ to create a counterfactual image $I^{*}$.
If the entity to intervene corresponds to multiple bounding boxes (\eg, when $\tilde{S}$ is ``a group of people''), all related regions are masked.
Next, we employ the initial image captioning model to generate a counterfactual caption $S^{*}$. 
The counterfactual captions are used to model $S^*_{<j}$ and $s^*_j$ in the causal graphs (Figure~\ref{fig:causal_graph}), \ie, modeling the generated words for the counterfactual image $I^*$.
Note that the goal of $S^*$ is to facilitate the estimation of causal effects, rather than serving as ground-truth captions for counterfactual images.
Thus, obtaining it is easier compared to acquiring ground-truth labels for counterfactual images. 
Consequently, we have $(I,S,\tilde{S},I^{*},S^{*})$ prepared for dataset $\mathcal{D}$.

\subsection{Our Framework}
\label{sec:objectives}
We propose a framework by incorporating negative log-likelihood (NLL) loss $\mathcal{L}_{\text{NLL}}$ with TE or NDE regularization loss, \ie, $\mathcal{L}_{\text{TE}}$ or $\mathcal{L}_{\text{NDE}}$, which will be described later.
Formally, the vanilla negative log-likelihood (NLL) loss is as follows:
\begin{equation}
    \mathcal{L}_{\text{NLL}} = -\sum_{(I,S)\in \mathcal{D}}\sum_{i=1}^{L} \log f_\theta(s_{i} \mid I, S_{<i}),
    \label{formula:nll_loss}
\end{equation}
where $f_{\theta}(\cdot)$ refers to the model that takes the image and preceding text sequence $S_{<i}$ as input and outputs a probability distribution on the vocabulary to generate the next token $s_{i}$, with parameters $\theta$.

We add the counterfactual regularization loss to allow the model to learn together with the NLL loss.
A hyperparameter $\alpha$ determines the weight of each loss, ensuring balanced optimization.
The final loss is denoted as:
\begin{equation}
\begin{aligned}
    &\mathcal{L}_{1} = \alpha \mathcal{L}_{\text{NLL}} + (1-\alpha)\mathcal{L}_{\text{TE}}, \\
    &\mathcal{L}_{2} = \alpha \mathcal{L}_{\text{NLL}} + (1-\alpha)\mathcal{L}_{\text{NDE}}.
    \label{formula:agg_loss}
\end{aligned}
\end{equation}

The whole optimization includes two stages: (1) training the model with vanilla NLL loss (Formula~\ref{formula:nll_loss}); (2) training the model with either $\mathcal{L}_{1}$ or $\mathcal{L}_{2}$ (Formula~\ref{formula:agg_loss}) using the constructed counterfactual images and their corresponding generated counterfactual captions.

\subsection{Total Effect Regularization}

When the region corresponding to ``black poodle'' is masked in the image (Figure~\ref{fig:framework}), we hope the model will significantly lower generation probabilities of the words ``black poodle'' to reduce hallucination.
To achieve this from a causal perspective, we maximize the total effect of changing $I$ to $I^*$ on the generation of ``black poodle'' (\ie, $\tilde{S}$), which is given in Formula~\ref{formula:te}.
Maximizing this total effect can be fulfilled by minimizing the following total effect (TE) loss:\looseness=-1 
\begin{equation}
\begin{aligned}
   \mathcal{L}_{\text{TE}} = -\sum_{(I,S)\in \mathcal{D}}&\sum_{j=1}^{L_{\tilde{S}}}\Big[\log f_{\theta}(\tilde{s}_{j}\mid I, S_{<p+j}) \\
    -&\frac{1}{L_{S^{*}}}\sum_{i=1}^{L_{S^{*}}}\log f_{\theta}(\tilde{s}_{j}\mid I^{*}, S_{<i}^{*})\Big],
    \label{formula:te_loss}
\end{aligned}
\end{equation}
where the first part corresponds to $Y_{I,M_I}$ in Formula~\ref{formula:te} and calculates the likelihood of generating $\tilde{S}$ (\eg, ``black poodle'') given the factual image $I$ and previously generated words, and the second part estimates $Y_{I^*,M_{I^*}}$ in Formula~\ref{formula:te} with the likelihood of generating $\tilde{S}$ at any position given the counterfactual image $I^{*}$ and preceding tokens $S^{*}_{<i}$ that are generated from $I^{*}$.
Here, $\tilde{s}_j$ denotes the $j$-th token of the entity to intervene $\tilde{S}$, of which the length is $L_{\tilde{S}}$.
$p$ represents the index of the first position where the entity to intervene $\tilde{S}$ appeared in the ground truth or factual caption (\eg, if ``black'' and ``poodle'' are the 9th and 10th words, then $p=9$). 
$L_{S^{*}}$ is the length of counterfactual caption $S^{*}$.
The specific occurrence position of the entity to intervene $\tilde{S}$ is explicit in the ground truth, while it may not necessarily appear in the counterfactual caption. 
Therefore, we need to estimate the probability of their occurrence using the average value.

In the example shown in Figure~\ref{fig:framework}, the word ``black poodle'' is the entity to intervene ($\tilde{S}$).
We estimate the first term by the probability of generating each token in ``black poodle'' at the position it appeared in the factual caption, \ie, using the preceding tokens ``A man in a yellow shirt training a''.
The second term is calculated by the average probabilities of generating each token in ``black poodle'' at any position in the counterfactual caption ``A man in a yellow shirt is stretching in a field.'', where the preceding tokens are those before each step.

\subsection{Natural Direct Effect Regularization}
To improve the visual perception ability of the model, another option is to maximize the natural direct effect (NDE) rather than the total effect (TE).
The natural direct effect is to measure the direct effect resulting from changes in the image.
As shown in Formula~\ref{formula:nde}, the first part is to calculate the likelihood of generating each token in the entity to intervene $\tilde{S}$ at any position, from the image $I$ and the preceding tokens $S^{*}_{<i}$ that have been generated from $I^*$.
Whereas, the second part is the likelihood of generating $\tilde{S}$ at any position from the counterfactual image $I^{*}$ and preceding tokens $S^{*}_{<i}$ that are generated from $I^{*}$, which is the same as the second part of TE.
Formally, we calculate NDE loss as:
\begin{equation}
\begin{aligned}
    \mathcal{L}_{\text{NDE}} = -\sum_{(I,S)\in \mathcal{D}}\sum_{j=1}^{L_{\tilde{S}}}&\Big[\frac{1}{L_{S^{*}}}\sum_{i=1}^{L_{S^{*}}}\Big(\log f_{\theta}(\tilde{s}_{j}\mid I, S_{<i}^{*}) \\ 
    &-\log f_{\theta}(\tilde{s}_{j}\mid I^{*}, S_{<i}^{*})\Big)\Big].
    \label{formula:nde_loss}
\end{aligned}
\end{equation}
The first component here appears simpler than that in the TE loss.
This is because, in the NDE loss, both the first and second components average the probabilities over any position in $S^{*}$, where it has a length of $L_{S^*}$.

\begin{table*}[t!]
\centering
\caption{
Evaluation results on counterfactual images with masks.
CH.$_{s}$ (CHAIR$_{s}$), P$_{@5}$ (Precision$_{@5}$), and nDCG$_{@5}$ are automatic measures for evaluating hallucination. Faith. (Faithfulness) and Overall denote results given by human judges. 
Methods are grouped by their shared backbone for clarity. 
The best result is highlighted in bold, while the second best is underlined. 
}
\label{table:hallu_res}
\setlength{\tabcolsep}{2.5mm}{
\begin{tabular}{l|cp{6mm}ccc|cp{6mm}ccc}
\toprule
\toprule
\multicolumn{1}{c|}{\multirow{2}{*}{\textbf{Methods}}} & \multicolumn{5}{c|}{\textbf{Flickr30k Entities}} & \multicolumn{5}{c}{\textbf{MSCOCO}} \\ \cmidrule(lr){2-11} 
\multicolumn{1}{c|}{} & \textbf{CH.$_{s}\downarrow$} & \textbf{P$_{@5}$} & \multicolumn{1}{c|}{\textbf{nDCG$_{@5}$}} & \textbf{Faith.} & \textbf{Overall} & \textbf{CH.$_{s}\downarrow$} & \textbf{P$_{@5}$} & \multicolumn{1}{c|}{\textbf{nDCG$_{@5}$}} & \textbf{Faith.} & \textbf{Overall} \\ \cmidrule(lr){1-11}

\textbf{ClipCap} & 20.45 & 80.08 & \multicolumn{1}{c|}{79.97} & 0.320 & 0.447 & 64.05 & 36.02 & \multicolumn{1}{c|}{36.01} & 0.687 & 0.713 \\
\quad+ObjL~\citep{biten2022let} & 21.18 & 79.16 & \multicolumn{1}{c|}{79.07} & 0.353 & 0.453 & 64.64 & 35.62 & \multicolumn{1}{c|}{35.58} & 0.653 & 0.727 \\
\quad+ObjMLM~\citep{dai2022plausible} & 25.37 & 75.07 & \multicolumn{1}{c|}{74.97} & 0.140 & 0.313 & 70.07 & 29.89 & \multicolumn{1}{c|}{29.91} & 0.533 & 0.547 \\
\quad+TE (ours) & \ul{19.78} & \ul{80.48} & \multicolumn{1}{c|}{\ul{80.39}} & \ul{0.373} & \ul{0.467} & \ul{63.58} & \ul{36.32} & \multicolumn{1}{c|}{\ul{36.35}} & \ul{0.753} & \ul{0.807} \\
\quad+NDE (ours) & \textbf{19.64} & \textbf{80.53} & \multicolumn{1}{c|}{\textbf{80.51}} & \textbf{0.400} & \textbf{0.493} & \textbf{63.04} & \textbf{36.55} & \multicolumn{1}{c|}{\textbf{36.65}} & \textbf{0.760} & \textbf{0.820} \\ \cmidrule(lr){1-11}

\textbf{BLIP} & 12.14 & 88.00 & \multicolumn{1}{c|}{88.00} & 0.740 & 0.793 & 33.70 & 66.17 & \multicolumn{1}{c|}{66.22} & 1.167 & 1.200 \\
\quad+ObjL~\citep{biten2022let} & 10.61 & 89.17 & \multicolumn{1}{c|}{89.19} & 0.613 & 0.767 & 33.07 & 67.12 & \multicolumn{1}{c|}{67.08} & 1.187 & 1.107 \\
\quad+ObjMLM~\citep{dai2022plausible} & \ul{10.11} & 89.31 & \multicolumn{1}{c|}{89.45} & 0.687 & 0.787 & 33.90 & 65.67 & \multicolumn{1}{c|}{65.77} & 1.113 & 1.180 \\
\quad+TE (ours) & 10.23 & \ul{89.63} & \multicolumn{1}{c|}{\ul{89.68}} & \ul{0.767} & \ul{0.827} & \ul{31.10} & \ul{68.49} & \multicolumn{1}{c|}{\ul{68.58}} & \ul{1.213} & \ul{1.267} \\
\quad+NDE (ours) & \textbf{9.53} & \textbf{89.83} & \multicolumn{1}{c|}{\textbf{89.93}} & \textbf{0.873} & \textbf{0.913} & \textbf{30.43} & \textbf{69.24} & \multicolumn{1}{c|}{\textbf{69.33}} & \textbf{1.247} & \textbf{1.273} \\ \cmidrule(lr){1-11}

\textbf{BLIP2} & 8.01 & 91.95 & \multicolumn{1}{c|}{91.96} & 0.807 & 0.847 & 30.28 & 69.91 & \multicolumn{1}{c|}{69.88} & 1.227 & 1.233 \\
\quad+ObjL~\citep{biten2022let} & 8.02 & 91.90 & \multicolumn{1}{c|}{91.96} & 0.847 & \ul{0.887} & 30.26 & 70.19 & \multicolumn{1}{c|}{70.13} & 1.133 & 0.947 \\
\quad+ObjMLM~\citep{dai2022plausible} & 8.12 & 92.00 & \multicolumn{1}{c|}{92.01} & 0.800 & 0.867 & 34.84 & 65.23 & \multicolumn{1}{c|}{65.19} & 1.140 & 1.100 \\
\quad+TE (ours) & \ul{7.61} & \ul{92.09} & \multicolumn{1}{c|}{\ul{92.14}} & \textbf{0.867} & \textbf{0.913} & \ul{29.60} & \ul{70.54} & \multicolumn{1}{c|}{\ul{70.49}} & \ul{1.340} & \ul{1.273} \\
\quad+NDE (ours) & \textbf{7.51} & \textbf{92.21} & \multicolumn{1}{c|}{\textbf{92.24}} & \ul{0.860} & 0.880 & \textbf{29.26} & \textbf{70.70} & \multicolumn{1}{c|}{\textbf{70.68}} & \textbf{1.353} & \textbf{1.280} \\ 
\bottomrule
\bottomrule
\end{tabular}}
\end{table*}

Figure~\ref{fig:framework}~(b) presents an example. The first term is estimated by the probability of generating each token in ``black poodle'' at any position in the counterfactual caption ``A man in a yellow shirt is stretching in a field.'', but with the factual image $I$ as input.
The second term is again the average probabilities of generating each token in ``black poodle'' at any position in the counterfactual caption with the counterfactual image $I^{*}$ as input.
By maximizing the NDE effect, the direct influence of the image is enhanced, thereby the model is more inclined to see the image and generate the correct next token, rather than to guess it.

\section{Experiments}

In this section, we conduct extensive experiments to evaluate the capability of our model for alleviating object hallucination, reducing biases in training data, and interpreting the correspondence between captions and image regions.

\subsection{Experiment Setup}
\subsubsection{Datasets} \quad
To evaluate the effectiveness of our model, we construct counterfactual images and captions as mentioned in Section~\ref{sec:construct_cf_data}.
We choose Flickr30k Entities~\cite{plummer2015flickr30k} and MSCOCO~\cite{lin2014microsoft} as our datasets, which have high-quality image annotations for constructing masked counterfactual images.

\noindent\textbf{Flickr30k Entities} (Flickr) is built upon the existing Flickr30k dataset~\cite{young2014image} that contains 31,783 images.
The dataset provides 244k coreference chains and 276k manually annotated bounding boxes within the images.
We use the original split of this dataset.
Entities that occur more than once within a caption are removed to avoid confusion.
After pre-processing, the final dataset consists of 29k/1k/1k samples for training/validation/test, respectively.

\noindent\textbf{MSCOCO} (COCO) consists of more than 328k images with annotated objects, phrases, and relationships.
We adopt the Karpathy split of the MSCOCO dataset and coreference relationships in the annotations are utilized to establish correspondences between the image regions and phrases (entities) to intervene in the captions.

Both datasets are composed of diverse phrase categories, where the Flickr dataset covers over 1,000 categories, while the COCO dataset is more concentrated on 80 categories.

\subsubsection{Backbones and Baselines} \quad
Our proposed counterfactual regularization losses are model-agnostic and can be applied to various models.
We conduct experiments on three backbones: ClipCap~\cite{mokady2021clipcap}, BLIP~\cite{li2022blip} and BLIP2~\cite{li2023blip2}, which respectively serve as representative models for decoder-only, encoder-decoder, and multimodal large language model architectures.
In addition to the above three image captioning backbones, we compare our methods with another two baselines that aim to alleviate the object hallucination in image captioning:
(1) \textbf{ObjL}~\cite{biten2022let} utilizes object labels as training augmentation to diminish models’ object bias on hallucination;
(2) \textbf{ObjMLM}~\cite{dai2022plausible} conducts a whole object mask to mitigate object hallucination in masked language modeling.
Both of them can also be applied to various backbones for a fair comparison.

\begin{table*}[t!]
\centering
\caption{Evaluation results on factual images without masks. BLEU-4, ROUGE-L, and CIDEr are automatic measures for evaluating generation quality. Faith. (Faithfulness) and Overall denote content accuracy and overall caption quality given by human judges. The best result is highlighted in bold, while the second best is underlined.}
\label{table:gen_res}
\setlength{\tabcolsep}{2mm}{
\begin{tabular}{l|ccccc|ccccc}
\toprule
\toprule
\multicolumn{1}{c|}{\multirow{2}{*}{\textbf{Methods}}} & \multicolumn{5}{c|}{\textbf{Flickr30k Entities}} & \multicolumn{5}{c}{\textbf{MSCOCO}} \\ \cmidrule(lr){2-11} 
 & \textbf{BLEU-4} & \textbf{ROUGE-L} & \multicolumn{1}{c|}{\textbf{CIDEr\ \ }} & \textbf{\ Faith.} & \textbf{Overall\ } & \textbf{BLEU-4} & \textbf{ROUGE-L} & \multicolumn{1}{c|}{\textbf{CIDEr\ \ }} & \textbf{\ Faith.} & \textbf{Overall\ } \\ \cmidrule(lr){1-11}
 
\textbf{ClipCap} & \ul{23.38} & \ul{48.33} & \multicolumn{1}{c|}{57.09} & 0.947 & 0.793 & \ul{28.79} & 52.75 & \multicolumn{1}{c|}{\ul{126.92}} & 1.360 & 1.253 \\
\quad+ObjL~\citep{biten2022let} & 23.58 & 48.17 & \multicolumn{1}{c|}{58.13} & 0.927 & 0.787 & 26.67 & 49.40 & \multicolumn{1}{c|}{115.15} & 1.373 & 1.173 \\
\quad+ObjMLM~\citep{dai2022plausible} & 17.01 & 43.64 & \multicolumn{1}{c|}{32.42} & 0.793 & 0.813 & 19.25 & 46.76 & \multicolumn{1}{c|}{73.82} & 1.040 & 1.013 \\
\quad+TE (ours) & \textbf{24.03} & \textbf{48.92} & \multicolumn{1}{c|}{\textbf{59.08}} & \textbf{0.973} & \textbf{0.860} & \textbf{28.94} & \textbf{52.98} & \multicolumn{1}{c|}{\textbf{127.27}} & \ul{1.413} & \textbf{1.300} \\
\quad+NDE (ours) & 23.32 & 48.05 & \multicolumn{1}{c|}{\ul{58.69}} & \ul{0.960} & \ul{0.840} & 28.77 & \ul{52.89} & \multicolumn{1}{c|}{125.70} & \textbf{1.433} & \ul{1.280} \\ \cmidrule(lr){1-11}

\textbf{BLIP} & \ul{37.14} & 56.67 & \multicolumn{1}{c|}{\ul{95.40}} & 1.433 & 1.260 & 34.63 & \textbf{56.83} & \multicolumn{1}{c|}{\ul{153.39}} & 1.873 & 1.593 \\
\quad+ObjL~\citep{biten2022let} & 36.93 & 56.37 & \multicolumn{1}{c|}{92.72} & 1.420 & 1.147 & 33.00 & 56.66 & \multicolumn{1}{c|}{148.00} & 1.840 & 1.300 \\
\quad+ObjMLM~\citep{dai2022plausible} & 35.61 & 56.56 & \multicolumn{1}{c|}{94.88} & 1.420 & 1.253 & 31.71 & 65.55 & \multicolumn{1}{c|}{133.67} & 1.713 & 1.540 \\
\quad+TE (ours) &  \textbf{37.28} &  \textbf{56.77} & \multicolumn{1}{c|}{\ul{95.40}} & \ul{1.447} & \ul{1.300} & \ul{34.65} & \ul{56.82} & \multicolumn{1}{c|}{153.37} & \ul{1.900} & \ul{1.620} \\
\quad+NDE (ours) & 37.00 & \ul{56.68} & \multicolumn{1}{c|}{ \textbf{95.45}} & \textbf{1.473} & \textbf{1.307} & \textbf{34.66} & \textbf{56.83} & \multicolumn{1}{c|}{\textbf{153.73}} & \textbf{1.920} & \textbf{1.640} \\ \cmidrule(lr){1-11}

\textbf{BLIP2} &  \ul{37.61} & \ul{58.11} & \multicolumn{1}{c|}{ \ul{103.41}} & 1.473 & 1.373 & 34.72 & 58.13 & \multicolumn{1}{c|}{154.17} & 1.880 & 1.553 \\
\quad+ObjL~\citep{biten2022let} & 34.30 & 56.61 & \multicolumn{1}{c|}{93.43} & 1.473 & 1.240 & 29.81 & 55.13 & \multicolumn{1}{c|}{135.42} & 1.820 & 1.340 \\
\quad+ObjMLM~\citep{dai2022plausible} & 36.04 & 56.94 & \multicolumn{1}{c|}{94.76} & 1.480 & 1.220 & 34.06 & 57.39 & \multicolumn{1}{c|}{151.51} & 1.800 & 1.453 \\
\quad+TE (ours) & \textbf{37.64} & \textbf{58.24} & \multicolumn{1}{c|}{\textbf{103.68}} & \ul{1.520} & \ul{1.393} & \ul{34.80} & \ul{58.24} & \multicolumn{1}{c|}{\ul{154.77}} & \ul{1.933} & \textbf{1.647} \\
\quad+NDE (ours) & 37.56 &  \ul{58.11} & \multicolumn{1}{c|}{102.63} & \textbf{1.533} & \textbf{1.427} & \textbf{34.88} & \textbf{58.34} & \multicolumn{1}{c|}{\textbf{155.14}} & \textbf{1.947} & \ul{1.613} \\ 
\bottomrule
\bottomrule
\end{tabular}}
\end{table*}

\subsubsection{Evaluation Methodology} \quad
Compared to baselines, our methods are expected to significantly reduce object hallucination on counterfactual test sets while maintaining the generation ability on factual test sets (it is not trivial due to different distributions between training and test sets).
We employ both automatic and human evaluation in our experiments for convincing conclusions.

\noindent \textbf{Automatic Evaluation:} 
We evaluate hallucination by using:

\textbf{$\bullet$ CHAIR$_{s}$}~\cite{rohrbach2018object}: It measures whether models generate a masked phrase, \ie, phrases whose corresponding regions have been masked in the counterfactual image:
\begin{equation}
    \begin{aligned}
        \text{CHAIR}_{s} = \frac{|\{\text{captions with hallucinated objects}\}|}{|\{\text{all captions}\}|},
    \end{aligned}
\end{equation}
where a lower CHAIR$_{s}$ score indicates a reduced presence of hallucination or increased faithfulness.

$\bullet$ Ranking-based Metrics: We generate five candidate captions with the highest probability of being generated for a given counterfactual image.
Captions without the masked phrase are considered positive, while otherwise are negative.
\textbf{Precision$_{@5}$} and \textbf{nDCG$_{@5}$} are employed\footnote{\url{https://github.com/microsoft/rankerEval}} to assess the object hallucination in fine-grained.

We adopt \textbf{BLEU}~\cite{papineni2002bleu}, \textbf{ROUGE-L}~\cite{lin2004rouge}, and \textbf{CIDEr}~\cite{vedantam2015cider} to measure the quality of generated captions on factual image test sets. We do not evaluate the generation quality of counterfactual images using automatic evaluation due to the lack of ground-truth captions. To compensate for this, the quality of captions generated for counterfactual images is analyzed by using human evaluation.

\noindent \textbf{Human Evaluation:} To verify whether the automatic measurements are consistent with human experiences, we further conduct a user study.
First, we randomly sample 50 factual images from the Flickr and 50 from the COCO dataset. We then create counterfactual images for the 100 factual images and collect top-generated captions from all methods for both factual and counterfactual images.
We conduct human evaluations on the 100 factual images and 100 counterfactual images.
For each image, we shuffle the generated captions and make the methods anonymous when presented with an image to ensure a fair comparison.
Three human assessors majored in English with the age range from 23 to 25, are hired to rate the captions on a 3-level Likert scale from 0 to 2 in two aspects:

$\bullet$ \textbf{Faithfulness} measures the degree to which a caption accurately represents the content of the image;

$\bullet$ \textbf{Overall} means the overall quality of a caption.

Finally, we calculate the Fleiss' Kappa among their assessments which results in 0.43, meaning a moderate level of agreement. We use their average values as the results.

\begin{figure*}[t!]
\includegraphics[width=1.\linewidth]{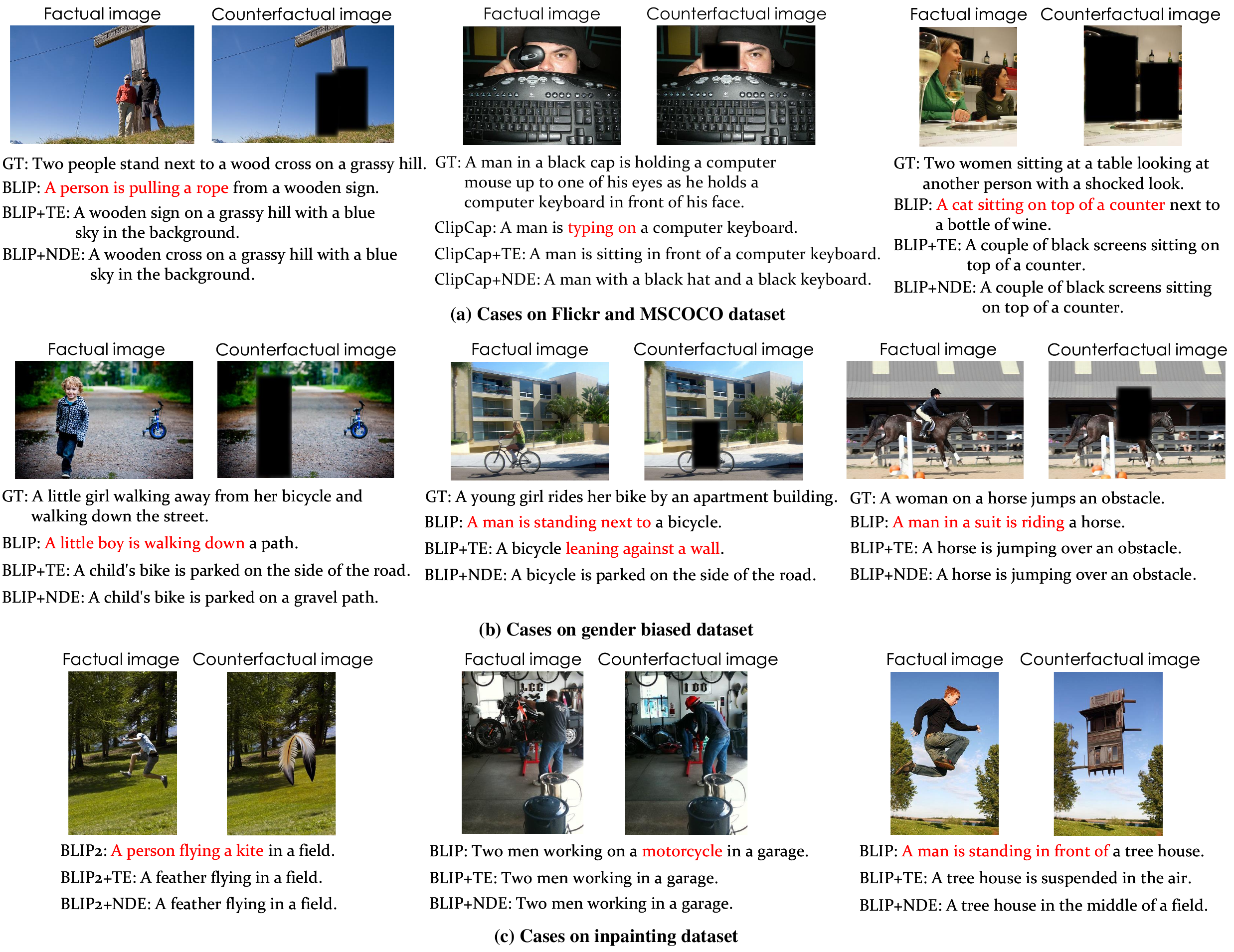}
\caption{Examples of generated captions by different methods on some masked or inpainted counterfactual images. Phrases highlighted in red are hallucinations that do not exist in the counterfactual image.}
\label{fig:case_study}
\end{figure*}

\subsection{Evaluation Results on Counterfactual Images}

We first compare our proposed models with all baselines on counterfactual test sets in terms of both hallucination and overall generation quality.
The results are shown in Table~\ref{table:hallu_res}, where all methods with the same backbone are grouped for clarity.

\textbf{Automatic evaluation}.
In terms of the automatic metrics of measuring object hallucination in Table~\ref{table:hallu_res} (CH.$_s$, P$_{@5}$, nDCG$_{@5}$), our proposed counterfactually regularized methods consistently exhibit superior performance over all baselines on both datasets, demonstrating their effectiveness in mitigating object hallucination.
Notably, our NDE regularization performs better than the TE ones.
It indicates that maximizing the direct effect of image content on the generated tokens helps build a more precise alignment between visual regions and their corresponding entity phrases.
Baselines ObjL and ObjMLM do not always alleviate hallucination effectively, \eg, they exhibit more hallucinations on the ClipCap backbone on the two datasets.
In contrast, our methods that regularize the causal effect consistently reduce hallucination in terms of different backbones, datasets, and measures. 
This demonstrates the effectiveness of adopting a causal perspective when handling hallucinations.
Further experiments confirm that different decoding strategies will not affect this (see Section~\ref{sec:appendix_decode} in the appendix).

\textbf{Human evaluation}.
Human evaluation results in Table~\ref{table:hallu_res} show that our methods perform the best regarding both reducing hallucination (Faith.) and overall generation quality (Overall). 
Moreover, NDE surpasses TE more frequently, and both proposed methods consistently outperform the baselines ObjL and ObjMLM. 
Overall, the human evaluation results are consistent with the automatic evaluation results in terms of hallucination and additionally reveal the good generation quality of our methods.

\subsection{Evaluation Results on Factual Images}
We compare all methods on factual test images to investigate 1) whether our regularization methods compromise any generation capability and 2) whether our method can reduce hallucination on factual images.
As shown in Table~\ref{table:gen_res},
our proposed methods achieve comparable or superior performance to all the baselines on both datasets in terms of automatic metrics for evaluating generation quality (BLEU, ROUGE-L, CIDEr). 
Human evaluation results also show that our methods can consistently outperform all baselines in reducing hallucination (Faithfulness) and increasing overall generation quality (Overall).
This indicates that our methods can significantly reduce hallucinations on counterfactual images without scarification in generation performance on factual images.

\begin{table}[t]
    \centering
    \caption{Error rate (out of 2,034 samples) of predicting female as male on two test sets. We show the number of samples with errors in parentheses. }
    \setlength{\tabcolsep}{2mm}{
    \begin{tabular}{l|cc}
    \toprule
    \textbf{Error Rate} & \textbf{Factual Image} & \textbf{Counterfactual Image} \\ \cmidrule(lr){1-3}
    BLIP & \ul{13.91\% (283)} & 38.25\% (778) \\
    BLIP+TE & 13.96\% (284) & \textbf{34.12\% (694)} \\
    BLIP+NDE & \textbf{13.27\% (270)} & \ul{34.27\% (697)} \\ \bottomrule
    \end{tabular}}
    \label{table:biased_error}
\end{table}

\subsection{Evaluation over Biased Datasets}

It would be interesting to investigate whether the proposed methods perform better when the test data has a biased distribution of some entities from training data.
Therefore, we construct a biased dataset from Flickr30k Entities. 
First, we do statistics of all the captions and find that 9,893 captions contain male-related words, such as ``man/men'' and ``boy/boys'', and 5,963 captions contain female-related words.
We then reconstruct a training set consisting of 8,942 male, 1,962 female, and 14,838 other captions, where the ratio of male to female is about 5:1.
We reverse the ratio to reconstruct the test set, which consists of 481 males, 2,034 females, and 500 other captions.
The validation set is constructed with a similar size and recipe as the test set.
Finally, we train a BLIP model and evaluate its performance on the biased test set.

When examining entities related to gender only, the error rates are presented in Table~\ref{table:biased_error}. 
The results indicate that the BLIP+NDE model outperforms BLIP in terms of lower error rates for both factual and counterfactual images.
It implies that models with our proposed methods are more robust in handling biased datasets.

\begin{figure}[t!]
\includegraphics[width=1.\linewidth]{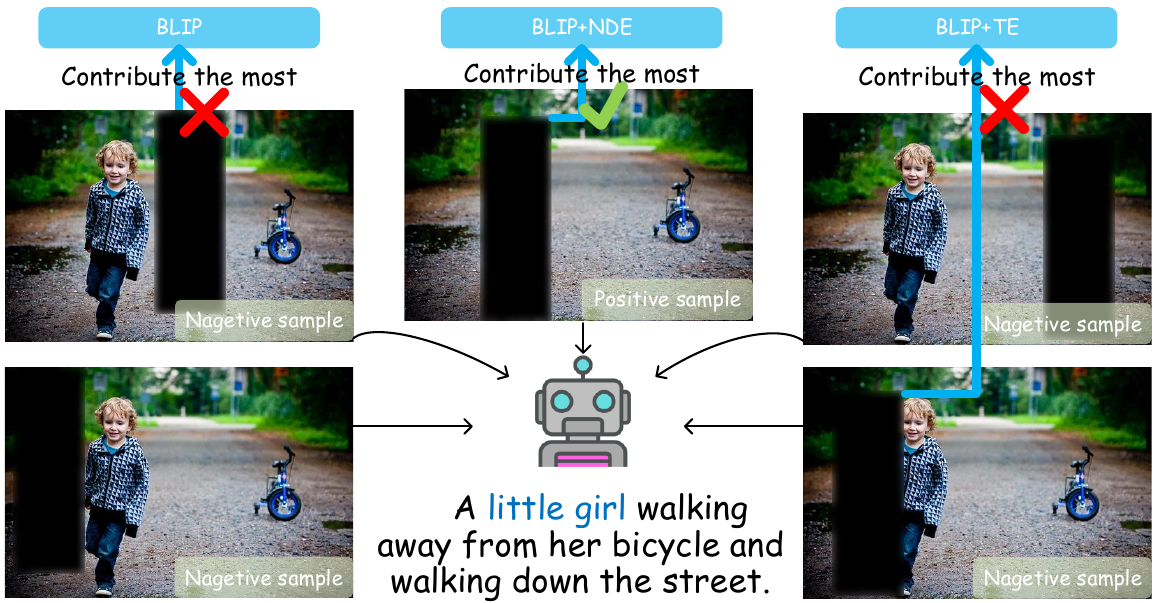}
\caption{Illustration of interpretability evaluation.}
\label{fig:interpretability}
\end{figure}

\subsection{Quantitative Analysis}

We present some examples of the generated captions on counterfactual images in Figure~\ref{fig:case_study}~(a), and more in supplementary pages.
Overall, our methods perform better in understanding counterfactual images, avoiding generating captions containing masked information.
Instead, they describe what is indeed presented in the images, such as ``a wooden cross'' and ``a couple of black screens''.
Conversely, the baseline model without counterfactual regularization often guesses incorrectly.
We also present some examples of generated captions for counterfactual images in the biased dataset in Figure~\ref{fig:case_study}~(b). 
The baseline model often incorrectly guesses a ``man'' or ``boy'' behind the mask, whereas our models describe other objects that are present in the image, such as ``a child's bicycle'' and ``a horse''.\looseness=-1

We further utilize a Latent Diffusion Model~\cite{ldm2022cvpr} to inpaint the masked region with a counterfactual object.
As shown in Figure~\ref{fig:case_study}~(c), an intriguing observation is that the baseline model occasionally hallucinates ``a person'' in the inpainted image, despite the absence of any human presence in the image.
This may be caused by the shortcut connections it learned from the training data, where our methods can robustly avoid this and correctly describe ``a feather'' or ``a tree house'' that can be seen in the inpainted images.

\begin{figure}[t!]
\includegraphics[width=1\linewidth]{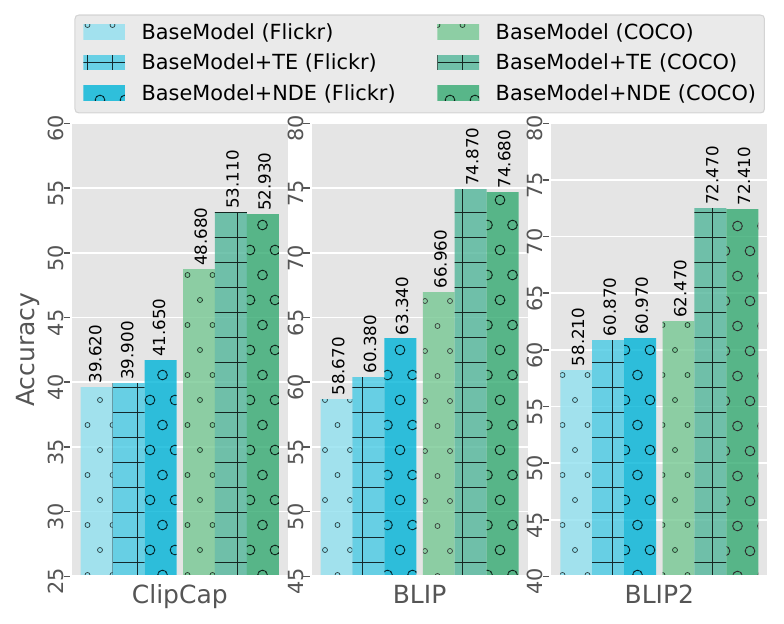}
\caption{The interpretability performance of different models by identifying the correct masked counterfactual image.}
\label{fig:loss_acc}
\end{figure}

\subsection{Evaluation of Interpretability}
Understanding the model's behavior is crucial for interpretability~\cite{zhang2024distillation,yang2024foundation}.
In this experiment, we compare different image captioning models based on interpretability.
An interpretable model should generate a noun phrase by utilizing its corresponding region. 
For example, when generating the phrase ``little girl'', the region containing a little girl should contribute the most to the model generation compared with other regions.
The contribution of a region is measured by using an efficient and effective explanation method CXPlain~\cite{schwab2019cxplain}. 
Specifically, for each noun phrase, we first identify its corresponding region (positive sample) and then randomly generate four incorrect regions with the same size (negative samples).
We then rank the five regions using CXPlain~\cite{schwab2019cxplain}, which assigns higher contribution scores to regions whose removal causes a larger change in the model's loss.
A model is considered interpretable if the positive sample receives the highest contribution score. 
Figure~\ref{fig:interpretability} shows an example, where the correct region (the region with a little girl) contributes the most to generating the phrase ``little girl'' for the Blip+NDE model, demonstrating the model's interpretability.
In contrast, for other models (Blip and Blip+TE), the correct region does not have the highest contribution to generating ``little girl''.

To present the average results over our test set, we use accuracy, defined as the percentage of cases where the positive sample has the highest contribution.
Figure~\ref{fig:loss_acc} shows that the TE method consistently performs better than the backbone model without regularization.
The NDE method significantly outperforms the TE method across backbones on Flickr, while both perform comparably on MSCOCO.
This suggests that our proposed counterfactual regularization effectively enhances interpretability, with the NDE method being the most effective.

\section{Conclusion}

This paper proposes using counterfactual causal effects to model the relationship between vision and language. 
We employ two counterfactual regularization methods based on the concepts of total effect (TE) and natural direct effect (NDE) to improve image captioning models.
Experimental results consistently show the superiority of our methods over baselines in terms of alleviating hallucination across different backbones and datasets. 
The NDE method performs the best in generating faithful captions for counterfactual images and accurately interpreting the most relevant image regions corresponding to a phrase in a caption.
In the future, we plan to integrate the counterfactual regularization methods into more complicated multimodal generation scenarios with both image and text as input, such as visual question answering and multimodal dialogue.

\section*{Limitations}
Hallucination and interpretability are important research areas across multiple disciplines.
Although we have explored the phenomenon of object hallucination and demonstrated the effectiveness of our methods in reducing it, a comprehensive understanding of the causal mechanisms underlying the appearance of hallucinations remains elusive and presents a more challenging problem.
Another limitation is that some issues may be related to limited data and model size.
While larger models have the potential to reduce errors, we were unable to conduct experimental verification due to insufficient GPU resources.

\begin{acks}
This work is supported by the National Key R\&D Program of China (2023YFF0905402), Beijing Natural Science Foundation (L233008), National Natural Science Foundation of China (No. 62276268), Migu Culture Technology Co., Ltd, Beijing Zhidemai Technology Co., Ltd, and Tencent AI Lab Rhino-Bird Focused Research Program.
We acknowledge the anonymous reviewers for their helpful comments.
\end{acks}

\bibliographystyle{ACM-Reference-Format}
\balance
\bibliography{main}

\clearpage
\appendix

\twocolumn[
\begin{center}
  \Huge\sffamily\bfseries 
  Supplementary Materials
\end{center}
\vspace{10pt}
]

\section{More Experimental Results}

\subsection{Evaluation on Decoding Algorithms}
\label{sec:appendix_decode}
Our method penalizes the occurrence probability of specific tokens, which may raise concerns regarding reliance on the generation algorithm.
To address this, we conducted experiments employing different decoding algorithms, \eg, greedy search, top-K sampling, and nucleus sampling, besides the beam search strategy mentioned before.
In our experiments, we set beam=5 for beam search, K=10 for top-K sampling, and p=0.8 for nucleus sampling.
As presented in Table~\ref{table:gen_algorithm}, the results on CHAIR$_{s}$ demonstrate that our methods consistently outperform the baselines, regardless of the decoding algorithm.
It is worth noting that our primary focus lies in examining the distinctions among various methods within the same decoding algorithm, rather than emphasizing the differences between different decoding algorithms.
Thus we adopt beam search as our general setting in our previous sections. 
This observation highlights the robustness of our approaches across various decoding algorithms, further proving their effectiveness.

\begin{table}[htbp!]
\small
\centering
\caption{Evaluation results of different decoding algorithms on CHAIR$_{s}$ values on Flickr30k Entities and MSCOCO.}
\label{table:gen_algorithm}
\resizebox{0.47\textwidth}{!}{
\setlength{\tabcolsep}{0.5mm}{
\begin{tabular}{p{17mm}cccccccc}
\toprule
\multirow{2}{*}{\textbf{Methods}} & \multicolumn{4}{c}{\textbf{Flickr30k Entities}} & \multicolumn{4}{c}{\textbf{MSCOCO}} \\ \cmidrule(lr){2-9} 
  & \textbf{Beam} & \textbf{Greedy} & \textbf{TopK} & \textbf{Nucleus} & \textbf{Beam} & \textbf{Greedy} & \textbf{TopK}  & \textbf{Nucleus} \\ \cmidrule(lr){1-9}
\textbf{ClipCap}& 20.45 & 19.82 & 15.06 & 15.55 & 64.05 & 66.43 & 56.00 & 62.45 \\
\quad+ObjL & 21.18 & 20.34 & 16.28 & 16.49 & 64.64 & 65.48 & 56.80 & 61.81 \\
\quad+ObjMLM & 25.37 & 24.54 & 19.12 & 19.89 & 70.07 & 74.20 & 65.56 & 67.99 \\
\quad+TE (ours) & \ul{19.78} & \ul{19.29} & \ul{14.92} & \textbf{14.57} & \ul{63.58} & \ul{64.97} & \ul{54.73} & \ul{59.09} \\
\quad+NDE (ours) & \textbf{19.64} & \textbf{19.05} & \textbf{14.60} & \ul{14.99} & \textbf{63.04} & \textbf{64.22} & \textbf{52.71} & \textbf{58.27} \\ \cmidrule(lr){1-9}

\textbf{BLIP} & 12.14 & 12.01 & 9.51 & 8.51 & 33.70 & 35.04 & 30.17 & 31.04 \\
\quad+ObjL & 10.61 & 11.52 & 9.41 & 7.51 & 33.07 & 32.56 & 27.86 & 30.08 \\
\quad+ObjMLM & \ul{10.11} & 11.01 & 7.71 & 7.29 & 33.90 & 36.45 & 30.70 & 31.26 \\
\quad+TE (ours) & 10.23 & \ul{10.92} & \ul{7.60} & \ul{6.81} & \ul{31.10} & \ul{32.02} & \ul{27.71} & \ul{28.40} \\
\quad+NDE (ours) & \textbf{9.53} & \textbf{10.51} & \textbf{7.51} & \textbf{6.80} & \textbf{30.43} & \textbf{31.04} & \textbf{26.97} & \textbf{27.86} \\ \cmidrule(lr){1-9}

\textbf{BLIP2} & 8.01 & \ul{7.60} & 7.81 & 6.52 & 30.28 & \ul{28.46} & 22.58 & 26.21 \\
\quad+ObjL & 8.02 & 7.74 & 7.08 & 7.46 & 30.26 & 29.45 & 22.20 & 25.51 \\
\quad+ObjMLM & 8.12 & 7.64 & 7.57 & 7.67 & 34.84 & 32.00 & 25.57 & 29.52 \\
\quad+TE (ours) & \ul{7.61} & \textbf{7.39} & \ul{6.24} & \textbf{6.10} & \ul{29.60} & \textbf{28.08} & \ul{22.14} & \textbf{23.64} \\
\quad+NDE (ours) & \textbf{7.51} & \textbf{7.39} & \textbf{6.21} & \ul{6.17} & \textbf{29.26} & \textbf{28.08} & \textbf{21.99} & \ul{24.22} \\ \bottomrule
\end{tabular}}}
\end{table}

\subsection{Results on More Inpainted Counterfactual Samples}
We have tested our method on more realistic counterfactual images (see Figure~\ref{fig:case_study}~(c) in the paper).
However, to verify our method in more high-quality counterfactual images, we experiment on the COCO-Counterfactuals (COCO-CF) dataset~\cite{le2024coco}, which automatically generates counterfactual examples based on MSCOCO using text-to-image diffusion models.
We adopt two test settings on the COCO-CF test set using models trained on COCO and COCO-CF, respectively.
As shown in Table~\ref{table:test_on_cf}, our method TE and NDE outperform the baselines on COCO-CF, demonstrating our superiority and generality.
Nevertheless, it is worth noting that COCO-CF is still a dataset with limited quality samples and lacks true counterfactuals with reasonable object associations.
How to further verify the capability of image captioning models on more realistic and higher quality counterfactual data requires more effort in the future.

\subsection{Validation on Larger Backbones}
We have validated the effectiveness of our method on an LLM, which is BLIP2 that has more than 3B parameters (ViT-L and OPT-2.7B, see Table~\ref{table:hallu_res} and~\ref{table:gen_res}).
We further conduct an experiment, leveraging a larger LLM, OPT-6.7B, to replace the decoder in BLIP2 (OPT-2.7B). 
As shown in Table~\ref{table:llm_6b7}, our methods consistently outperform baselines across all metrics.

\begin{table}[h]
\centering
\caption{Results of testing on MSCOCO with BLIP2-6.7B.}
\label{table:llm_6b7}
\setlength{\tabcolsep}{1.2mm}{
\begin{tabular}{c|ccccc}
\toprule
\textbf{Metrics} & \textbf{BLIP2-6.7B} & \textbf{+ObjL} & \textbf{+ObjMLM} & \textbf{+TE} & \textbf{+NDE} \\ \cmidrule(lr){1-6}

\textbf{CH.$_{s}\downarrow$} & 29.77 & 29.26 & 29.26 & \textbf{28.31} & \ul{28.46} \\
\textbf{P$_{@5}$} & 69.63 & 70.59 & 70.50 & \ul{71.17} & \textbf{71.45} \\
\textbf{nDCG$_{@5}$} & 69.73 & 70.64 & 70.58 & \ul{71.28} & \textbf{71.50} \\ \cmidrule(lr){1-6}

\textbf{BLEU-4} & 34.01 & 32.51 & 31.60 & \ul{34.16} & \textbf{34.22} \\ 
\textbf{ROUGE-L} & 57.87 & 56.60 & 56.22 & \ul{58.01} & \textbf{58.06} \\ 
\textbf{CIDEr} & 152.33 & 142.74 & 138.32 & \ul{153.04} & \textbf{153.67} \\ 
\bottomrule
\end{tabular}}
\end{table}

\subsection{Impact of $\alpha$}
\begin{figure}[h]
\includegraphics[width=1\linewidth]{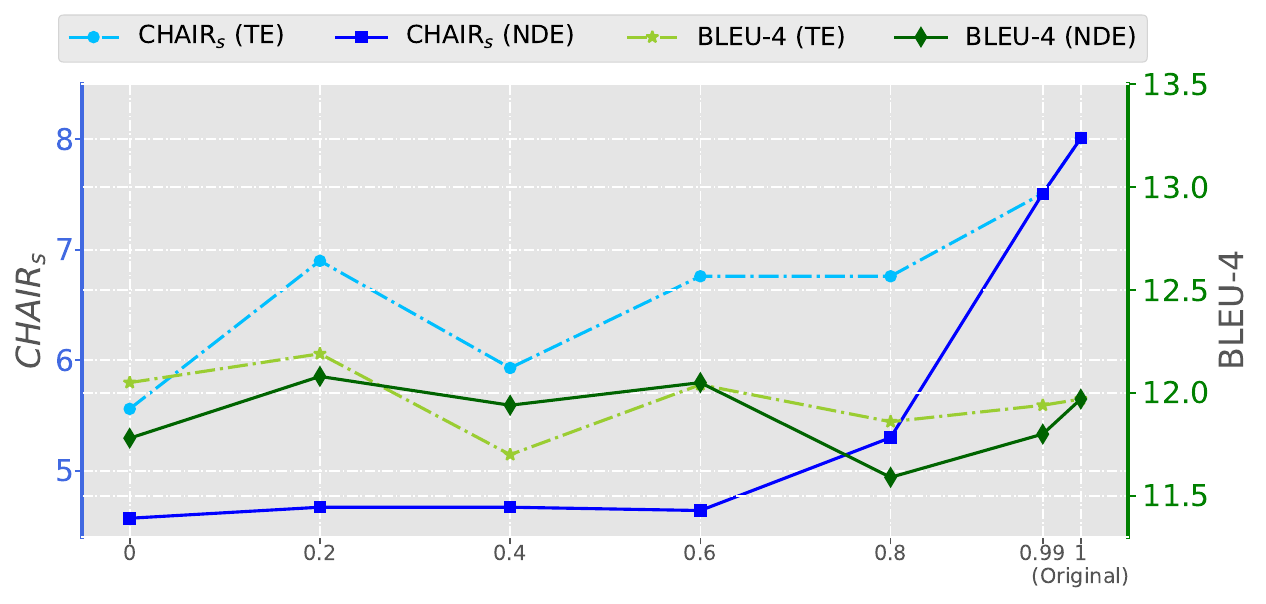}
\caption{The variation line chart of CHAIR$_{s}$ and BLEU-4 of BLIP2 when $\alpha$ changes on TE and NDE. The value of CHAIR$_{s}$ shows a clear change trend, while the value of BLEU-4 fluctuates insignificantly. Best viewed in color.}
\label{fig:alpha}
\end{figure}

We further investigate the impact of the hyperparameter $\alpha$ on producing object hallucination.
BLIP2 is adopted as the backbone and the results are depicted in Figure~\ref{fig:alpha}.
Generally speaking, CHAIR$_{s}$ gradually rises as $\alpha$ increases.
This makes sense since either our TE or NDE methods serve as a regularization.
The less the regularization, the worse the result.
It experiences substantial alterations while the parameter alpha ranges between 0.8 and 1.
However, when $\alpha$ gradually converges to 1, the model degenerates into a vanilla training process.
In addition, we find no significant drops in model generation performance on factual images as $\alpha$ decreases.
This evidence substantiates the assertion that our approach maintains the model's performance intact in factual scenarios.

\begin{table*}[t!]
\caption{The results on COCO-CF (of two settings where models are trained on COCO and COCO-CF, respectively).}
\centering
\label{table:test_on_cf}
\begin{tabular}{c|ccccc|ccccc}
\toprule
\multicolumn{1}{c|}{\multirow{2}{*}{\textbf{Metrics}}} & \multicolumn{5}{c|}{\textbf{Trained on COCO}} & \multicolumn{5}{c}{\textbf{Trained on COCO-CF}} \\ \cmidrule(lr){2-11} 
\multicolumn{1}{c|}{} & \textbf{\ BLIP2} & \textbf{+ObjL} & \textbf{+ObjMLM} & \textbf{+TE} & \textbf{+NDE\ } & \textbf{\ BLIP2} & \textbf{+ObjL} & \textbf{+ObjMLM} & \textbf{+TE} & \textbf{+NDE} \\ \cmidrule(lr){1-11}

\textbf{CH.$_{s}\downarrow$} & 8.17 & 9.30 & 9.78 & \ul{7.92} & \textbf{7.87} & 7.96 & 7.98 & 7.70 & \ul{7.45} & \textbf{7.33} \\
\textbf{P$_{@5}$} & 91.74 & 90.56 & 90.10 & \ul{92.07} & \textbf{92.08} & 92.03 & 91.75 & 91.96 & \textbf{92.47} & \ul{92.42} \\
\textbf{nDCG$_{@5}$} & 91.74 & 90.57 & 90.13 & \ul{92.07} & \textbf{92.09} & 92.02 & 91.82 & 92.05 & \textbf{92.48} & \ul{92.45} \\ \cmidrule(lr){1-11}

\textbf{BLEU-4} & 13.81 & 11.76 & 12.90 & \textbf{14.18} & \ul{14.17} & 16.22 & 14.43 & 14.47 & \textbf{16.42} & \ul{16.40} \\
\textbf{ROUGE-L} & 43.23 & 41.31 & 41.83 & \ul{43.30} & \textbf{43.33} & 46.22 & 45.09 & 44.57 & \textbf{46.36} & \ul{46.35} \\
\textbf{CIDEr} & 152.39 & 131.35 & 138.81 & \ul{154.03} & \textbf{154.17} & 180.04 & 166.82 & 164.26 & \textbf{180.35} & \ul{180.27} \\
\bottomrule
\end{tabular}
\end{table*}

\subsection{Results on Cross-domain Scenarios}
The impact of distribution differences between test and training data may affect the generalization ability of the model.
Thus we test the different methods in a cross-domain setting to provide more comprehensive results.
Specifically, we evaluate the models trained on COCO on the Flickr test set.
As shown in Table~\ref{table:cross_domain}, our methods perform better than all baselines, indicating our superior generalization ability.

\begin{table}[h]
    \centering
    \caption{Testing BLIP2 (trained on COCO) on Flickr.}
    \setlength{\tabcolsep}{1.2mm}{
    \begin{tabular}{c|ccccc}
    \toprule
    \textbf{Metrics} & \textbf{BLIP2} & \textbf{+ObjL} & \textbf{+ObjMLM} & \textbf{+TE} & \textbf{+NDE} \\ \cmidrule(lr){1-6}
    \textbf{CH.$_{s}\downarrow$} & 6.37 & 6.20 & 6.23 & \ul{6.06} & \textbf{6.02} \\
    \textbf{P$_{@5}$} & 93.31 & 93.69 & 93.77 & \ul{93.92} & \textbf{93.98} \\
    \textbf{nDCG$_{@5}$} & 93.37 & 93.70 & 93.76 & \ul{93.92} & \textbf{93.98} \\
    \bottomrule
    \end{tabular}}
    \label{table:cross_domain}
\end{table}

\section{Implementation Details}
\label{sec:appendix_imple}
As previously elucidated, our optimization contains two stages.
In stage one, ClipCap and BLIP are trained on Flickr30k Entities and MSCOCO for 10 epochs with a learning rate at 5e-5/1e-5, and the batch size is set to 128/32, respectively.
As for BLIP2, the learning rate is respectively set to 1e-6/7e-6 for Flickr and MSCOCO, and it is trained for 5 epochs with a batch size of 64.
In stage two, the TE/NDE loss is added for 2 epochs until the aggregate loss on validation converges, with the hyperparameter $\alpha$ set to 1-1e-3/1-1e-8 on Flickr30k Entities and 1-1e-4/1-3e-5 on MSCOCO for ClipCap, while for BLIP it is 1-9e-3/1-9e-3 on Flickr30k Entities and 1-5e-4/1-6e-4 on MSCOCO, and 1-1e-4/1-1e-2 on Flickr30k Entities and 1-1e-4 on MSCOCO for BLIP2, respectively.
For BLIP2, we adopt ViT-L and OPT-2.7b as the visual encoder and the language model, which are frozen during training.
We use beam search for all backbones with a beam size of 5 and a maximum length of 20 during inference.

\section{Further Analysis and Discussions}

\subsection{Clarification on Generating CF Captions}
When generating counterfactual (CF) captions, the initial model is a base image captioning model trained by fine-tuning with factual images.
We do not need to ensure that counterfactual captions are free of hallucinations.
Even if the CF captions contain hallucinations, the model is not guided to generate such hallucinations.
The goal of CF captions is to facilitate the estimation of causal effects, not being used as a ground-truth caption for training counterfactual images. 
Instead, the CF captions are used as the preceding tokens to compute the generation probability of a target token that introduces hallucinations in the counterfactual scenario. 
We minimize this probability to reduce hallucinations.
The higher the probability of the target entity, the more it will be subtracted in the TE/NDE loss calculation.
The CF caption, essentially a sample of the model's probability distribution, ensures consistency with the model's behavior for accurate token probability assessment.

\subsection{Advantages and Disadvantages of TE/NDE}
Although both of our methods benefit from the concepts of causal modeling, they have different advantages and disadvantages from each other.
NDE focuses on hallucination alleviation, while TE brings better generation performance.
This is because NDE models only the direct influence of the image on word tokens, and TE additionally models the influence of the preceding word tokens. 
Thus, NDE describes visual information more accurately, and TE generates text with rich and smooth semantics, making them suitable for different scenarios.
Their differences may account for the different performances displayed in Table~\ref{table:hallu_res} and~\ref{table:gen_res}.

\subsection{More Cases}
\begin{figure*}[t!]
\includegraphics[width=1.\linewidth]{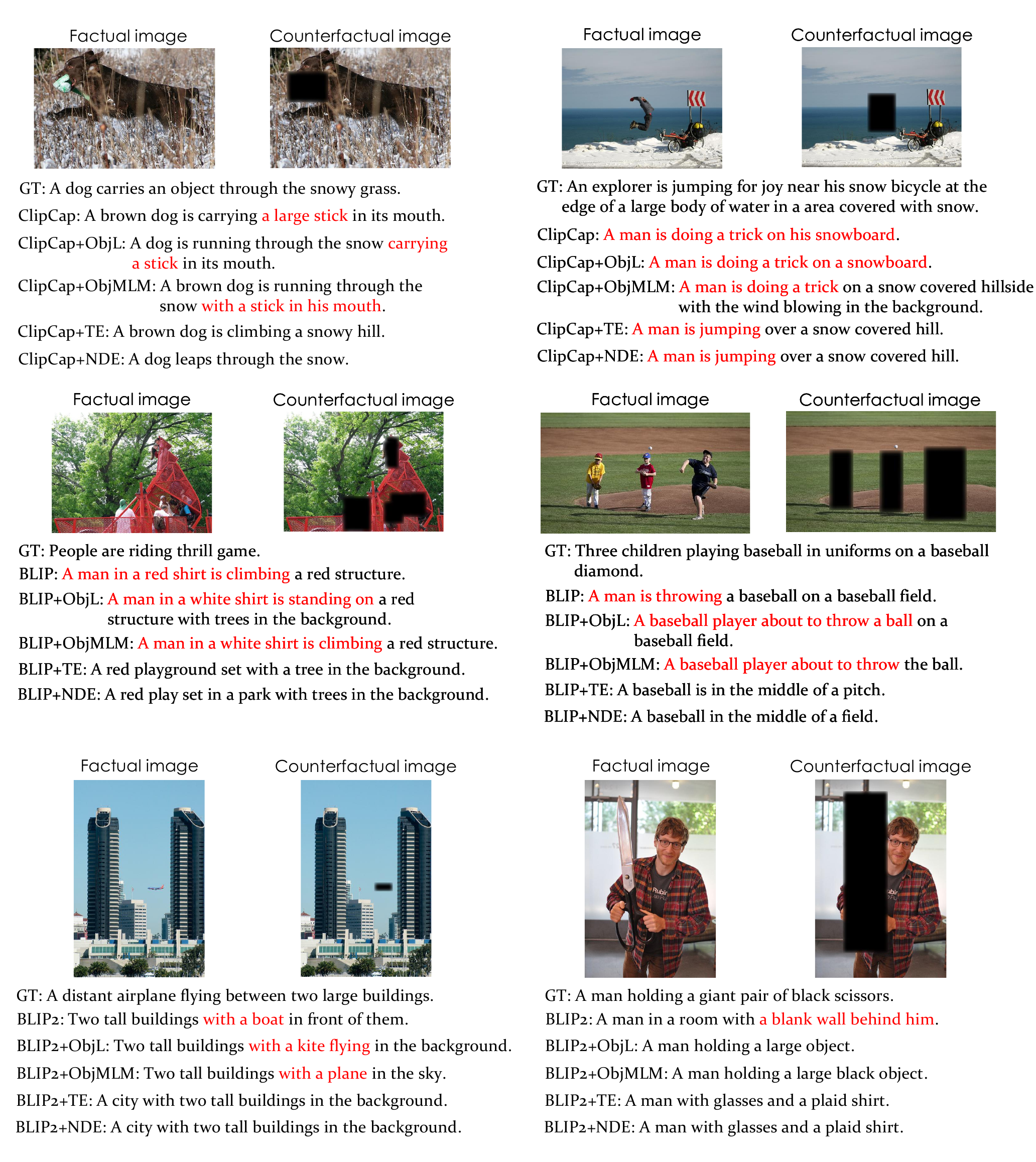}
\caption{Some examples of generated captions by various methods for some masked counterfactual images. Phrases highlighted in red are hallucinations that do not exist in the counterfactual image.}
\label{Xsuppl:case1}
\end{figure*}

\begin{figure*}[t!]
\includegraphics[width=1.\linewidth]{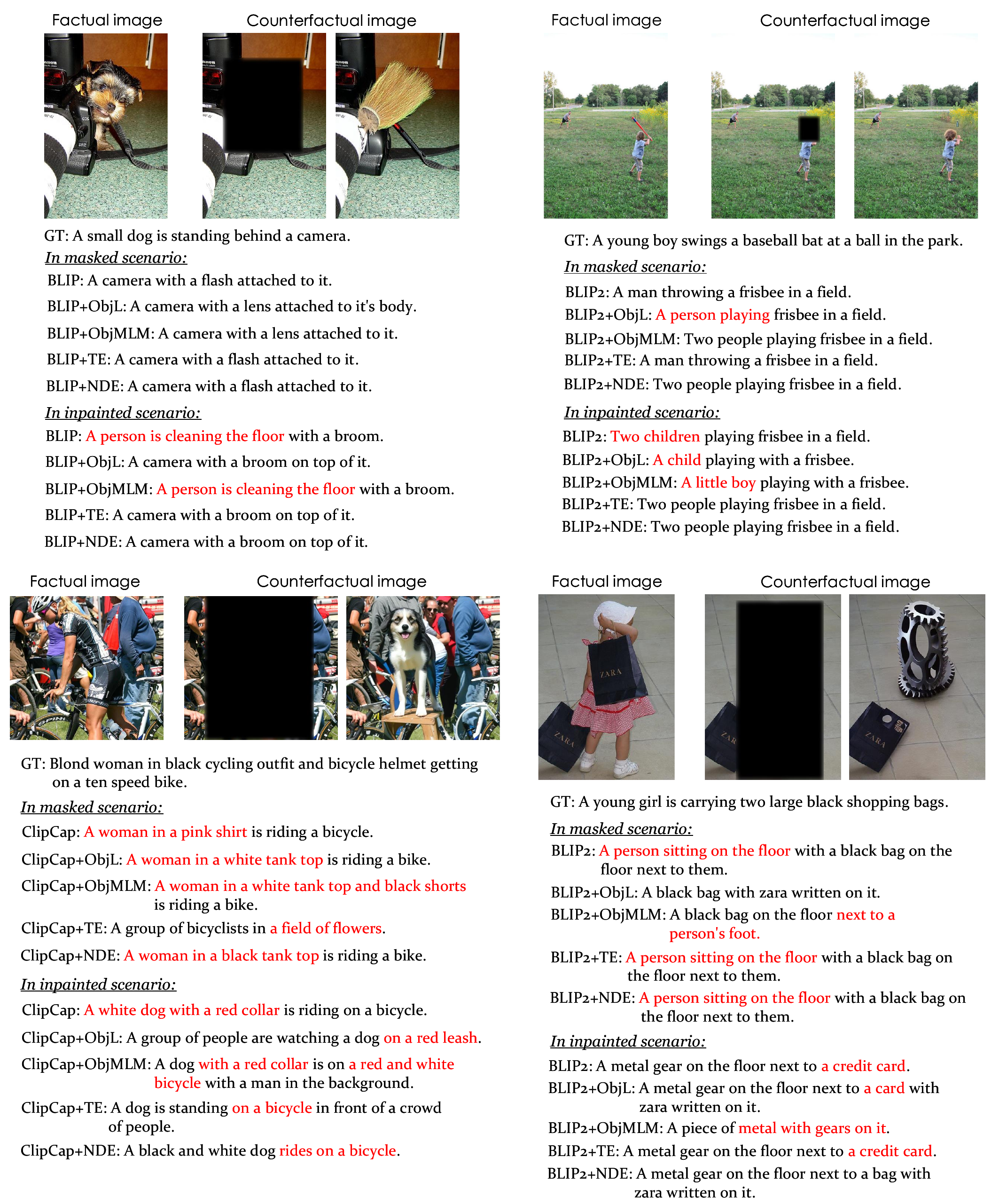}
\caption{Some examples of generated captions by various methods for some masked and inpainted counterfactual images. Phrases highlighted in red are hallucinations that do not exist in the counterfactual image.}
\label{Xsuppl:case2}
\end{figure*}

We present more cases of masked counterfactual images and inpainted counterfactual images across different methods, displayed in Figure~\ref{Xsuppl:case1} and~\ref{Xsuppl:case2}.
Through the comparison of various methodologies, our approach consistently generates image captions that exhibit greater fidelity to the underlying visual content.
By avoiding unreliable conjectures, our methods can successfully mitigate the occurrence of object hallucinations, thereby augmenting the robustness and reliability of various image captioning models.
Nevertheless, all methods may still encounter challenges in some complex situations.
A comprehensive analysis and subsequent improvements are necessary to enhance both reliability and validity in future investigations.

\end{document}